\documentclass[runningheads]{llncs}
\usepackage{graphicx}
\usepackage{amsmath,amssymb} 
\usepackage{array}
\usepackage{multirow}
\usepackage{color}
\usepackage{lipsum}
\usepackage{mwe}

\usepackage[width=122mm,left=12mm,paperwidth=146mm,height=193mm,top=12mm,paperheight=217mm]{geometry}

\begin{document}
\newcommand*\rot{\rotatebox{90}}
\pagestyle{headings}
\mainmatter

\title{Modeling Camera Effects to Improve Visual Learning from Synthetic Data} 

\titlerunning{Modeling Camera Effects}
\authorrunning{A. Carlson et al.}


\author{Alexandra Carlson$^{\star}$, Katherine A. Skinner\thanks{Authors contributed equally to this work.}, Ram Vasudevan, and Matthew Johnson-Roberson}
\institute{University of Michigan, Ann Arbor}

\maketitle

\begin{abstract}

Recent work has focused on generating synthetic imagery to increase the size and variability of training data for learning visual tasks in urban scenes. This includes increasing the occurrence of occlusions or varying environmental and weather effects. 
However, few have addressed modeling variation in the sensor domain. 
Sensor effects can degrade real images, limiting generalizability of network performance on visual tasks trained on synthetic data and tested in real environments. 
This paper proposes an efficient, automatic, physically-based augmentation pipeline to vary sensor effects -- chromatic aberration, blur, exposure, noise, and color temperature -- for synthetic imagery. 
In particular, this paper illustrates that augmenting synthetic training datasets with the proposed pipeline reduces the domain gap between synthetic and real domains for the task of object detection in urban driving scenes. 

\end{abstract}

\section{Introduction}

Deep learning has enabled impressive performance increases across a range of computer vision tasks. 
However, this performance improvement is largely dependent upon the size and variation of labeled training datasets that are available for a chosen task. For some tasks, benchmark datasets contain millions of hand-labeled images for the supervised training of deep neural networks (DNNs) ~\cite{krizhevsky2012imagenet,zhou2017places}.
Ideally, we could compile a large, comprehensive training set that is representative of all domains and is labelled for all visual tasks. However, it is expensive and time-consuming to both collect and label large amounts of training data, especially for more complex tasks like detection or pixelwise segmentation~\cite{Cityscapes}. Furthermore, it is practically impossible to gather a single real dataset that captures all of the variability that exists in the real world.

Two promising methods have been proposed to overcome the limitations of real data collection: graphics rendering engines and image augmentation pipelines. These approaches enable increased variability of scene features across an image set without requiring any additional manual data annotation. Recent work in rendering datasets has shown success in training DNNs with large amounts of highly photorealistc, synthetic data and testing on real data ~\cite{johnson2017driving,ros2016synthia2}. Pixel-wise labels for synthetic images can be generated automatically by rendering engines, greatly reducing the cost and effort it takes to create ground truth for different tasks. 
Recent work on image augmentation has focused on modeling environmental effects such as scene lighting, time of day, scene background, weather, and occlusions in training images as a way to increase the representation of these visual factors in training sets, thereby increasing robustness to these cases during test time~\cite{zhang2017image,veeravasarapu2017adversarially}. Another proposed augmentation approach is to increase the occurrence of objects of interest (such as cars or pedestrians) in images in order to provide more training examples of those objects in different scenes and spatial configurations ~\cite{alhaija_kittiaugmented,huangexpecting}.

\begin{figure}[t]
\begin{center}
\includegraphics[width=0.8\linewidth]{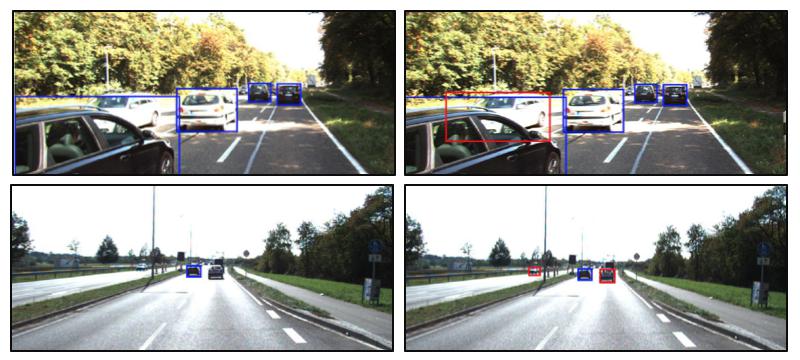}
\end{center}
\caption{Examples of object detection tested on KITTI for baseline unaugmented data (left) and for our proposed method (right). Blue boxes show correct detections; red boxes show detections missed by the baseline method but detected by our proposed approach for sensor-based image augmentation.}
\label{fig:motivate_failuremodes}
\end{figure}
However, even with varying spatial geometry and environmental factors in an image scene, there remain challenges to achieving robustness of task performance when transferring trained networks between synthetic and real image domains. 
To further understand the gaps between synthetic and real datasets, it is worthwhile to consider the failure modes of DNNs in visual learning tasks.
One factor that has been shown to contribute to degradation of performance and cross-dataset generalization for various benchmark datasets is sensor bias~\cite{andreopoulos2012sensor,song2015sun,dodge2016understanding,doersch2015unsupervised}. 
The interaction between the camera model and lighting in the environment can greatly influence the pixel-level artifacts, distortions, and dynamic range induced in each image ~\cite{grossberg2004modelingCRF,couzinie2013learning_modelingblur,foi2008practical}. 
Sensor effects, such as blur and overexposure, have been shown to decrease performance of object detection networks in urban driving scenes~\cite{failingtolearn}. Examples of failure modes caused by over exposure, manifesting as missed detections, are shown in Figure~\ref{fig:motivate_failuremodes}.
However, there still is an absence in the literature examining how to improve failure modes due to sensor effects for learned visual tasks in the wild.


In this work, we propose a novel framework for augmenting synthetic data with realistic sensor effects -- effectively randomizing the sensor domain for synthetic images. 
Our augmentation pipeline is based on sensor effects that occur in image formation and processing that can lead to loss of information and produce failure modes in learning frameworks -- chromatic aberration, blur, exposure, noise and color cast.
We show that our proposed method improves performance for object detection in urban driving scenes when trained on synthetic data and tested on real data, an example of which is shown in Figure~\ref{fig:motivate_failuremodes}. Our results demonstrate that sensor effects present in real images are important to consider for bridging the domain gap between real and simulated environments.

This paper is organized as follows: Section~\ref{sec:background} presents related background work; section~\ref{sec:techapproach} details the proposed image augmentation pipeline; section ~\ref{sec:expres} describes experiments and discusses results of these experiments and section~\ref{sec:concl} concludes the paper. Code for this paper can be found at\\ \verb|https://github.com/alexacarlson/SensorEffectAugmentation|.

\section{Related Work}
\label{sec:background}
\subsubsection{Domain randomization with synthetic data:} 
Rendering and gaming engines have been used to synthesize large, labelled datasets that contain a wide variety of environmental factors that could not be feasibly captured during real data collection~\cite{ros2016synthia,gaidon2016virtual}. Such factors include time of day, weather, and community architecture. Improvements to rendering engines have focused on matching the photorealism of the generated data to real images, which comes at a huge computational cost. 
Recent work on domain randomization seeks to bridge the reality gap by generating synthetic data with sufficient random variation over scene factors and rendering parameters such that the real data falls into this range of variation, even if rendered data does not appear photorealistic. 
Tobin et. al~\cite{tobin2017domain} focus on the task of object localization trained with synthetic data. They perform domain randomization over textures, occlusion levels, scene lighting, camera field of view, and uniform noise within the rendering engine, but their experiments are limited to highly simplistic toy scenes. 
Building on~\cite{tobin2017domain}, Tremblay et al~\cite{tremblay2018training} generate a synthetic dataset via domain randomization for object detection of real urban driving scenes. They randomize camera viewpoint, light source, object properties, and introduce flying distractors. Our work focuses on image augmentation outside of the rendering pipeline and could be applied in addition to domain randomization in the renderer.

\subsubsection{Augmentation with synthetic data:} 
Shrivastava et al. recently developed SimGAN, a generative adversarial network (GAN) to augment synthetic data to appear more realistic. They evaluated their method on the the tasks of gaze estimation and hand pose estimation~\cite{simgan}. Similarly, Sixt et al. proposed RenderGAN, a generative network that uses structured augmentation functions to augment synthetic images of markers attached to honeybees~\cite{sixt2016rendergan}. The augmented images are used to train a detection network to track the honeybees. Both of these approaches focus on image sets that are homogeneously structured and low resolution. We instead focus on the application of autonomous driving, which features highly varied, complex scenes and environmental conditions. 
\subsubsection{Traditional Augmentation Techniques:}
Standard geometric augmentations, such as rotation, translation, and mirroring, have become commonplace in deep learning for achieving invariance to spatial factors that are not relevant to the given task~\cite{hauberg2016dreaming}. 
Photometric augmentations aim to increase robustness to differing illumination color and intensity in a scene. These augmentations induce small changes in pixel intensities that do not produce loss of information in the image. A well known example is the PCA-based color shift introduced by Krizhevsky et al.~\cite{krizhevsky2012imagenet} to perform more realistic RGB color jittering. 
In contrast, our augmentations are modeled directly from real sensor effects and can induce large changes in the input data that mimics the loss of information that occurs in real data.
\subsubsection{Sensor effects in learning:} 
More generally, recent work has demonstrated that elements of the image formation and processing pipeline can have a large impact upon learned representation~\cite{Kanan2012,diamond2017dirty,dodge2016understanding}. Andreopoulos and Tsotsos demonstrate the sensitivities of popular vision algorithms under variable illumination, shutter speed, and gain~\cite{andreopoulos2012sensor}. Doersch et al. show there is dataset bias introduced by chromatic aberration in visual context prediction and object recognition tasks~\cite{doersch2015unsupervised}. They correct for chromatic aberration to eliminate this bias. Diamond et al. demonstrate that blur and noise degrade neural network performance on classification tasks~\cite{diamond2017dirty}. They propose an end-to-end denoising and deblurring neural network framework that operates directly on raw image data. 
Rather than correcting for the effects of the camera during image formation of real images, we propose to augment synthetic images to simulate these effects. As many of these effects can lead to loss of information, correcting for them is non-trivial and may result in the hallucination of visual information in the restored image. 

\section{Sensor-based Image Augmentation}
\label{sec:techapproach}
%
\begin{figure}[t]
\begin{center}
   \includegraphics[width=0.6\linewidth]{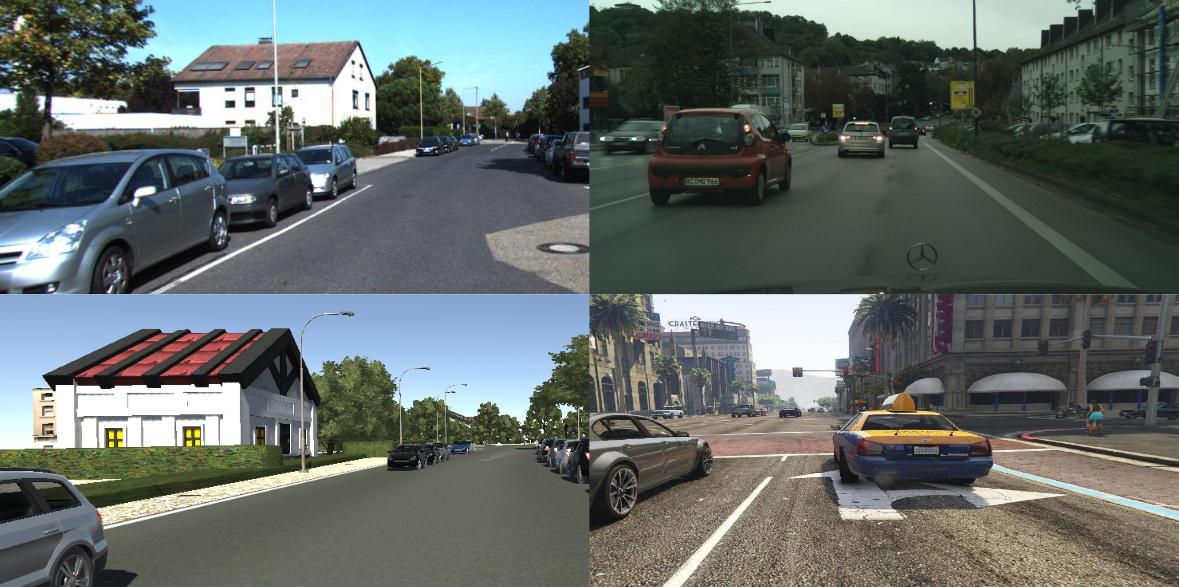}
\end{center}
   \caption{A comparison of images from the KITTI Benchmark dataset (upper left), Cityscapes dataset (upper right), Virtual KITTI (lower left) and Grand Theft Auto (lower right). Note that each dataset has differing color cast, brightness, and detail.}
\label{fig:datasets} 
\end{figure}
Figure~\ref{fig:datasets} shows a side-by-side comparison of two real benchmark vehicle datasets, KITTI~\cite{kittiobject,KITTI13} and Cityscapes~\cite{Cityscapes}, and two synthetic datasets, Virtual KITTI~\cite{gaidon2016virtual} and Grand Theft Auto ~\cite{Richter_2016_ECCV,johnson2017driving}. Both of the real datasets share many spatial and environmental visual features: both are captured during similar times of day, in similar weather conditions, and in cities regionally close together, with the camera located on a car pointing at the road. In spite of these similarities, images from these datasets are visibly different. This suggests that these two real datasets differ in their global pixel statistics.
Qualitatively, KITTI images feature more pronounced effects due to blur and over-exposure. Cityscapes has a distinct color cast compared to KITTI. 
Synthetic datasets such as Virtual KITTI and GTA have many spatial similarities with real benchmark datasets, but are still visually distinct from real data.
Our work aims to close the gap between real and synthetic data by modelling these sensor effects that can cause distinct visual differences between real world datasets.
%
\begin{figure*}
\begin{center}
\includegraphics[width=0.95\linewidth]{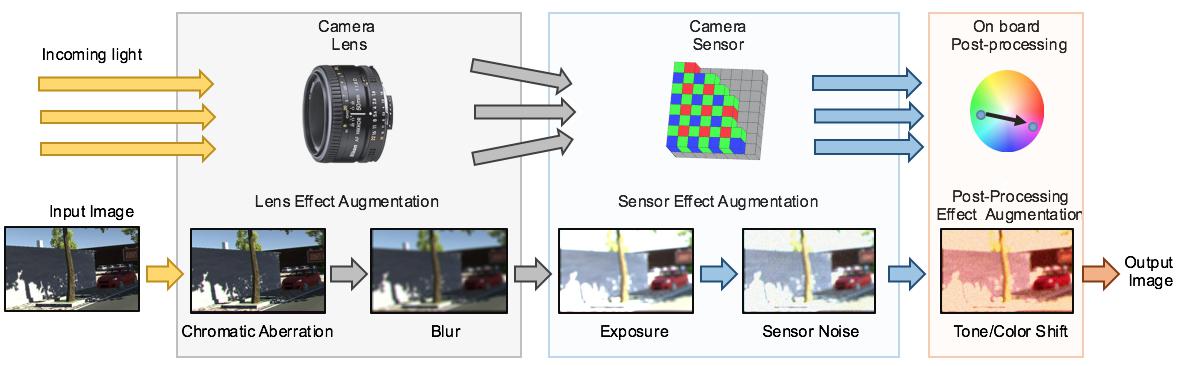}
\end{center}
\caption{A schematic of the image formation and processing pipeline used in this work.  A given image undergoes augmentations that approximate the same pixel-level effects that a camera would cause in an image.}
\label{fig:imagepipeline}
\end{figure*}
Figure~\ref{fig:imagepipeline} shows the architecture of the proposed sensor-based image augmentation pipeline.  
We consider a general camera framework, which transforms radiant light captured from the environment into an image~\cite{karaimer2016software}. There are several stages that comprise the process of image formation and post-processing steps, as shown in the first row of Figure~\ref{fig:imagepipeline}. 
The incoming light is first focused by the camera lens to be incident upon the camera sensor. Then the camera sensor transforms the incident light into RGB pixel intensity. On-board camera software manipulates the image (e.g., color space conversion and dynamic range compression) to produce the final output image. 
At each stage of the image formation pipeline, loss of information can occur to degrade the image. Lens effects can introduce visual distortions in an image, such as chromatic aberration and blur. Sensor effects can introduce over- or under-saturation depending on exposure, and high frequency pixel artifacts, based on characteristic sensor noise. Lastly, post-processing effects are implemented to shift the color cast to create a desirable output. 
Our image augmentation pipeline focuses on five total sensor effects augmentations to model loss of information that can occur at each stage during image formation and post-processing: chromatic aberration, blur, exposure, noise, and color shift. To model how these effects manifest in images in a camera, we implement the image processing pipeline as a composition of physically-based augmentation functions across these five effects, where lens effects are applied first, then sensor effects, and finally post-processing effects: 
\begin{align}
I_{aug.} = \phi_{color}(\phi_{noise}(\phi_{exposure}(\phi_{blur}(\phi_{chrom.ab.}(I)))))
\label{eq:aug-pipeline-eq}
\end{align}
Note that these chosen augmentation functions are not exhaustive, and are meant to approximate the camera image formation pipeline. Each augmentation function is described in detail in the following subsections.

\subsection{Chromatic Aberration} 
Chromatic aberration is a lens effect that causes color distortions, or fringes, along edges that separate dark and light regions within an image. There are two types of chromatic aberration, longitudinal and lateral, both of which can be modeled by geometrically warping the color channels with respect to one another~\cite{kang2007automatic_chromab}.
Longitudinal chromatic aberration occurs when different wavelengths of light converge on different points along the optical axis, effectively magnifying the RGB channels relative to one another. We model this aberration type by scaling the green color channel of an image by a value $S$. Lateral chromatic aberration occurs when different wavelengths of light converge to the different points within the image plane. We model this by applying translations $(t_{x},t_{y})$ to each of the color channels of an image. We combine these two effects into the following affine transformation, which is applied to each $(x,y)$ pixel location in a given color channel $C$ of the image:
\begin{align}
\begin{bmatrix}
    x^{\tiny{chrom.ab.}}_{C}  \\
    y^{\tiny{chrom.ab.}}_{C}  \\
    1 
\end{bmatrix} = 
\begin{bmatrix}
    S & 0 & t_{x} \\
    0 & S & t_{y} \\
    0 & 0 & 1
\end{bmatrix}
\begin{bmatrix}
    x_{C}  \\
    y_{C}  \\
    1 
\end{bmatrix}
\label{eq:chromabMats}
\end{align}

\subsection{Blur} 
While there are several types of blur that occur in image-based datasets, we focus on out-of-focus blur, which can be modeled using a Gaussian filter~\cite{cheong2015fast}:
\begin{align}
G = \frac{1}{2\pi\sigma^2}e^{-\frac{x^2+y^2}{2\sigma^2}}
\label{eq:blur}
\end{align}
\noindent where $x$ and $y$ are spatial coordinates of the filter and $\sigma$ is the standard deviation. The output image is given by:
\begin{align}
I_{blur} = I*G
\label{eq:blurimg}
\end{align}

\subsection{Exposure} 
To model exposure, we use the exposure density function developed in~\cite{exp1}~\cite{exp2}:
\begin{align}
I = f(S) = \frac{255}{1 + e^{-A \times S}}
\label{eq:exposure}
\end{align}
\noindent where $I$ is image intensity, $S$ indicates incoming light intensity, or exposure, and $A$ is a constant value for contrast. We use this model to re-expose an image as follows:
\begin{align}
S' = f^{-1}(I) +  \Delta S
\label{eq:exposureds}
\end{align}
\begin{align}
I_{exp} = f(S')
\label{eq:exposureimg}
\end{align}
\noindent We vary $\Delta S$ to model changing exposure, where a positive $\Delta S$ relates to increasing the exposure, which can lead to over-saturation, and a negative value indicates decreasing exposure.

\subsection{Noise}
The sources of image noise caused by elements of the sensor array can be modeled as either signal-dependent or signal-independent noise. Therefore, we use the Poisson-Gaussian noise model proposed in~\cite{foi2008practical}: 
\begin{align}
I_{noise}(x,y)=I(x,y)+\eta_{poiss}(I(x,y))+\eta_{gauss}
\label{eq:noise1}
\end{align}
where $I(x,y)$ is the ground truth image at pixel location $(x,y)$, $\eta_{poiss}$ is the signal-dependent Poisson noise, and $\eta_{gauss}$ is the signal-independent Gaussian noise.
We sample the noise for each pixel based upon its location in a \textit{GBRG} Bayer grid array assuming bilinear interpolation as the demosaicing function. 

\subsection{Post-processing}
In standard camera pipelines, post-processing techniques, such as white balancing or gamma transformation, are nonlinear color corrections performed on the image to compensate for the presence of different environmental illuminants. These post-processing methods are generally proprietary and cannot be easily characterized~\cite{grossberg2004modelingCRF}. 
We model these effects by performing translations in the CIELAB color space, also known as L*a*b* space, to remap the image tonality to a different range ~\cite{hunter1948accuracyLAB,annadurai2007fundamentals}. 
Given that our chosen datasets are all taken outdoors during the day, we assume a D65 illuminant in our L*a*b* color space conversion. 

\subsection{Generating Augmented Training Data}
\begin{figure}[t]
\begin{center}
\includegraphics[width=0.75\linewidth]{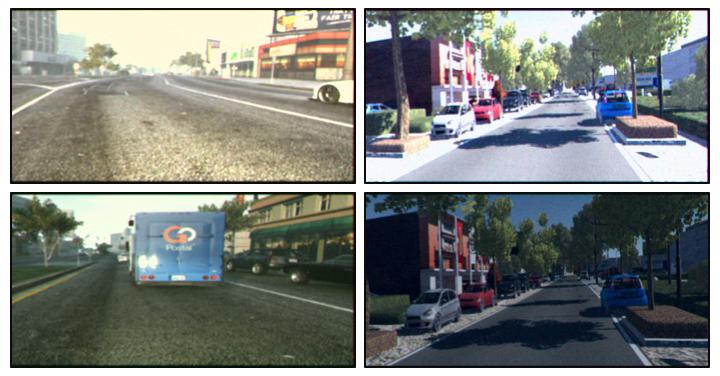}
\end{center}
\caption{Example augmentations of GTA (left column) and VKITTI (right column) using the proposed sensor effect augmentation pipeline. Each image has a randomly sampled level of blur, chromatic aberration, exposure, sensor noise, and color temperature shift applied to it in an effort to model the visual structure/information loss caused by cameras when capturing real images.}
\label{fig:ex_randalleffectaugs_real}
\end{figure}

The bounds on the sensor effect parameter regimes were chosen experimentally. The parameter selection process is discussed in more detail in Section~\ref{sec:expres}.
To augment an image, we first randomly sample from these visually realistic parameter ranges. Both the chosen parameters and the unaugmented image are then input to the augmentation pipeline, which outputs the image augmented with the camera effects determined by the chosen parameters. We augmented each image multiple times with different sets of randomly sampled parameters. Note that this augmentation method serves as a pre-processing step. 
Figure~\ref{fig:ex_randalleffectaugs_real} shows sample images augmented with individual sensor effects as well as our full proposed sensor-based image augmentation pipeline.
We use the original image labels as the labels for the augmented data. Pixel artifacts from cameras, like chromatic aberration and blur, make the object boundaries noisy. Thus, the original target labels are used to ensure that the network makes robust and accurate predictions in the presence of camera effects.



\section{Experiments}
\label{sec:expres}
%
We evaluate the proposed sensor-based image augmentation pipeline on the task of object detection on benchmark vehicle datasets to assess its effectiveness at bridging the synthetic to real domain gap. We apply our image augmentation pipeline to two benchmark synthetic vehicle datasets, each of which was rendered with different levels of photorealism.
The first, Virtual KITTI (VKITTI)~\cite{gaidon2016virtual}, features over 21000 images and is designed to models the spatial layout of KITTI with varying environmental factors such as weather and time of day. The second is Grand Theft Auto (GTA)~\cite{Richter_2016_ECCV,johnson2017driving}, which features 21000 images and is noted for its high quality and increased photorealism compared to VKITTI. 
To evaluate the proposed augmentation method for 2D object detection, we used Faster R-CNN as our base network~\cite{ren2015fasterRCNN}. Faster R-CNN achieves relatively high performance on the KITTI benchmark test dataset, and many state-of-the-art object detection networks that improve upon these results use Faster R-CNN as their base architecture. 
For all experiments, we apply sensor effect augmentation pipeline to all images in the given dataset, then train an object detection network on the combination of original unaugmented data and sensor effect augmented data. We ran experiments to determine the number of sensor effect augmentations per image, and determined that optimal performance was achieved by augmenting each image in each dataset one time. 
To determine the bounds of the sensor effect parameter ranges from which to sample, we augmented small datasets of 2975 images by randomly sampling from increasingly larger parameter bounds and chose the ranges for each sensor effect that yielded the highest performance as well as visually realistic images. We found that the same parameter regime yielded optimal performance for both synthetic datasets.
All of the trained networks are tested on a held out validation set of 1480 images from the KITTI training data and we report the Pascal VOC $AP50_{bbox}$ value for the car class. We also report the gain in $AP50_{bbox}$, which is the difference in performance relative to the baseline (unaugmented) dataset. 
We compare the performance of object detection networks trained on sensor-effect augmented data to object detection networks trained on unaugmented data as our baseline. 
For each dataset, we trained each Faster R-CNN network for 10 epochs using four Titan X Pascal GPUs in order to control for potential confounds between performance and training time.
While the focus of this paper is on synthetic image augmentation, we investigate sensor effect augmentation on reducing the domain gap between real data as supplementary material.

\subsection{Performance on baseline Object Detection Benchmarks}

\begin{table*}
\begin{center}
\caption{Object detection trained on synthetic data, tested on KITTI}
\label{fig:objdetsynth}
\begin{tabular}{c}

\begin{tabular}{lll}
\noalign{\smallskip}
\hline\noalign{\smallskip}
\multicolumn{3}{c}{Virtual KITTI}\\
\hline\noalign{\smallskip}
Training Set & $AP_{car}$ & Gain  \\
\noalign{\smallskip}
\hline
\noalign{\smallskip}
2975 Baseline & 54.60 & \textemdash \\
2975 Prop. Method & 61.88 &  $\uparrow$ 7.28 \\
\hline\noalign{\smallskip}
Full Baseline (21K) & 58.25 & \textemdash \\
Full Prop. Method & 62.52 & $\uparrow$ 4.27\\
\multicolumn{3}{c}{\textemdash}\\
\noalign{\smallskip}\hline
\end{tabular}

\\

\begin{tabular}{lll}
\noalign{\smallskip}
\hline\noalign{\smallskip}
\multicolumn{3}{c}{GTA}\\
\hline\noalign{\smallskip}
Training Set & $AP_{car}$ & Gain  \\
\noalign{\smallskip}
\hline
\noalign{\smallskip}
2975 Baseline  &  46.83 & \textemdash\\
2975 Prop. Method & 51.24 & $\uparrow$ 4.41\\
\hline\noalign{\smallskip}
Full Baseline (21K) &  49.80 & \textemdash\\
Full Baseline (50K) &  53.26 & \textemdash\\
Full Prop. Method  & 55.85 & $\uparrow$ 6.05\\
\noalign{\smallskip}\hline
\end{tabular}

\end{tabular}
\end{center}
\end{table*}
Table~\ref{fig:objdetsynth} shows results for FasterRCNN networks trained on  unaugmented synthetic data and sensor-effect augmented data for both VKITTI and GTA. 
Note that we provide experiments trained on the full training datasets, as well as experiments trained on subsets of 2975 images to allow comparison of performance across differently sized datasets.
Synthetic data augmented with the proposed method yields significant performance gains over the baseline (unaugmented) synthetic datasets. 
This is expected as, in general, rendering engines do not realistically model sensor effects such as noise, blur, and chromatic aberration as accurately as our proposed approach. Another important result for the synthetic datasets (both VKITTI and GTA), is that, by leveraging our approach, we are able to outperform the networks trained on over 20000 unaugmented images with a tiny subset of 2975 images augmented with using our approach. This means that not only can networks be trained faster but also when training with synthetic data, varying camera effects can outweigh the value of simply generating more data with varied spatial features.
The VKITTI baseline dataset tested on KITTI performs relatively well compared to GTA, even though GTA is a more photorealistic dataset. This can most likely be attributed to the similarity in spatial layout and image features between VKITTI and KITTI. With our proposed approach, VKITTI gives comparable performance to the network trained on the Cityscapes baseline, showing that synthetic data augmented with our proposed sensor-based image pipeline can perform comparably to real data for cross-dataset generalization. 

\subsection{Comparison to other Augmentation Techniques}
We ran experiments to compare our proposed method to photometric augmentation, specifically PCA-based color shift \cite{krizhevsky2012imagenet}, complex spatial/geometric augmentations, specifically elastic deformation \cite{ronneberger2015u}, standard additive gaussian noise augmentation, and a suite of standard spatial augmentations, specifically random rotations, scaling, translations, and cropping. We provide the results of training Faster-RCNN networks on the full VKITTI and GTA datasets augmented with the above methods in Table \ref{tab:otheraugs}. All networks were tested on the same held-out set of KITTI images as used in the previous object detection experiments.
Our results show that our proposed method drastically outperforms the other standard augmentation techniques, and that for certain synthetic data, spatial augmentations actually decrease performance on real data.
This suggests that the proposed sensor effect augmentations capture more salient visual structure than traditional, non-photorealistic augmentation methods. We hypothesize this is because the physically-based sensor augmentations better model the information loss and the resulting global pixel-statistics that occur in real images. For example, our proposed method uses LAB space color transformation to alter the color cast of an image, where as traditional approaches use RGB space. LAB space is device independent, so it results in a more accurate, physically-based augmentation than \cite{krizhevsky2012imagenet}.
\setlength{\tabcolsep}{1.0pt}
\begin{table*}
\begin{center}
\caption{We provide the results of training Faster-RCNN networks on GTA and Virtual KITTI augmented with various augmentation methods. All networks were tested on KITTI.}
\label{tab:otheraugs}
\begin{tabular}{lllllllllll}
\hline\noalign{\smallskip}
\multicolumn{10}{c}{Virtual KITTI}    \\
\hline
Augmentation Method       &&&&& $AP_{Car}$ &&&&& \footnotesize{Gain} \\
\noalign{\smallskip}
\hline
\noalign{\smallskip}
Baseline                                                          &&&&& 58.25          &&&&& \textemdash\\  
\textbf{Prop. Method}                                             &&&&& \textbf{62.52} &&&&& $\uparrow$ \textbf{4.27}\\  
\textit{Krishevsky et. al \cite{krizhevsky2012imagenet}}          &&&&& 59.09          &&&&& $\uparrow$ 0.84\\  
\textit{Ronneberger et. al \cite{ronneberger2015u}}               &&&&& 56.56          &&&&& $\downarrow$ 1.69\\ 
Additive Gaussian Noise                                           &&&&& 56.98          &&&&& $\downarrow$ 1.27 \\ 
\multirow{2}{10em}{\small{Random Rotation, Scale, Transl., Crop}} &&&&& 55.11          &&&&& $\downarrow$ 3.14\\
\noalign{\smallskip} \\
\hline
\hline
\noalign{\smallskip}%

\multicolumn{10}{c}{GTA}  \\
\hline
\noalign{\smallskip}
Augmentation Method           &&&&& $AP_{Car}$ &&&&& \footnotesize{Gain} \\
\noalign{\smallskip}
\hline
\noalign{\smallskip}
Baseline (21k)                                                    &&&&& 49.80          &&&&& \textemdash\\  
\textbf{Prop. Method (21k)}                                       &&&&& \textbf{55.85} &&&&& $\uparrow$ \textbf{6.05}\\  
\textit{Krishevsky et. al \cite{krizhevsky2012imagenet}}          &&&&& 51.62          &&&&& $\uparrow$ 1.88\\  
\textit{Ronneberger et. al \cite{ronneberger2015u}}               &&&&& 48.94          &&&&& $\uparrow$ 0.14\\  
Additive Gaussian Noise                                           &&&&& 52.01          &&&&& $\uparrow$ 2.21\\ 
\multirow{2}{10em}{\small{Random Rotation, Scale, Transl., Crop}} &&&&& 50.11          &&&&& $\uparrow$ 0.31\\
\noalign{\smallskip}\\
\hline
\end{tabular}
\end{center}
\end{table*}
\setlength{\tabcolsep}{1.4pt}

\subsection{Ablation Study}

\setlength{\tabcolsep}{1.0pt}
\begin{table*}
\begin{center}
\caption{Ablation study for object detection trained on synthetic data, tested on KITTI}
\label{fig:ablation_synthetic}
\begin{tabular}{lllll lllll llll l}
\hline\noalign{\smallskip}
\multicolumn{15}{c}{Virtual KITTI}    \\
\hline
\noalign{\smallskip}
Training Set &&&&& Augmentation Type &&&&& $AP_{car}$ &&&& Gain  \\
\noalign{\smallskip}
\hline
\noalign{\smallskip}
2975 Baseline     &&&&& \it{None}    &&&&& 54.60 &&&& \textemdash \\
2975 Prop. Method &&&&& Chrom. Ab.   &&&&& 61.08 &&&& $\uparrow$ 6.48 \\
2975 Prop. Method &&&&& Blur         &&&&& 59.72 &&&& $\uparrow$ 5.12 \\
2975 Prop. Method &&&&& Exposure     &&&&& 57.37 &&&& $\uparrow$ 2.77 \\
2975 Prop. Method &&&&& Sensor Noise &&&&& 58.60 &&&& $\uparrow$ 4.00 \\
2975 Prop. Method &&&&& Color Shift  &&&&& 58.59 &&&& $\uparrow$ 3.99 \\
\noalign{\smallskip}
\hline
\hline
\noalign{\smallskip}

\multicolumn{15}{c}{GTA}  \\
\hline
\noalign{\smallskip}
Training Set &&&&& Augmentation Type &&&&& $AP_{car}$ &&&& Gain  \\
\noalign{\smallskip}
\hline
\noalign{\smallskip}
2975 Baseline     &&&&& \it{None}    &&&&& 46.83 &&&& \textemdash \\
2975 Prop. Method &&&&& Chrom. Ab.   &&&&& 48.92 &&&& $\uparrow$ 2.09 \\
2975 Prop. Method &&&&& Blur         &&&&& 49.17 &&&& $\uparrow$ 2.34 \\
2975 Prop. Method &&&&& Exposure     &&&&& 47.95 &&&& $\uparrow$ 1.12 \\
2975 Prop. Method &&&&& Sensor Noise &&&&& 48.09 &&&& $\uparrow$ 1.26 \\
2975 Prop. Method &&&&& Color Shift  &&&&& 48.61 &&&& $\uparrow$ 1.78 \\
\hline
\end{tabular}
\end{center}
\end{table*}
\setlength{\tabcolsep}{1.4pt}

To evaluate the contribution of each sensor effect augmentation on performance, we used the proposed pipeline to generate datasets with only one type of sensor effect augmentation.
We trained Faster-RCNN on each of these datasets augmented with single augmentation functions, the results of which are given in Table~\ref{fig:ablation_synthetic}. 
Performance increases across all ablation experiments for training on synthetic data. This further validates our hypothesis that each of the sensor effects are important for closing the gap between synthetic and real data. 

\subsection{Failure Mode Analysis}
\begin{figure*}[t]
\begin{center}
\includegraphics[width=1.0\linewidth]{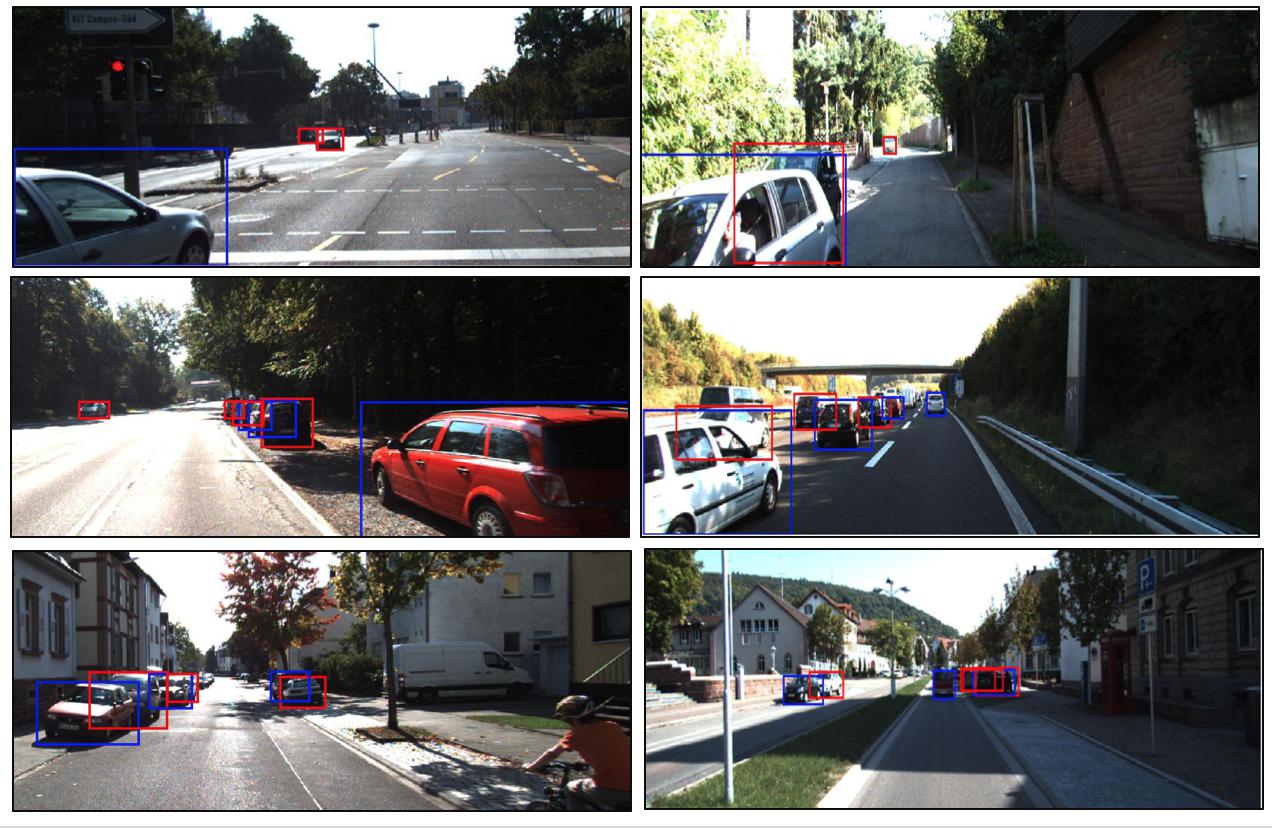}
\end{center}
\caption{Virtual KITTI examples are in the left column, GTA examples are in the right column. Blue boxes show correct detections; red boxes show detections missed by the FasterRCNN network trained on baseline, unaugmented image datasets but detected by FasterRCNNs trained on data augmented using our proposed approach for sensor-based image augmentation.}
\label{fig:failmode_synth}
\end{figure*}

Figure~\ref{fig:failmode_synth} shows the qualitative results of failure modes of FasterRCNN trained on each synthetic training dataset and tested on KITTI, where the blue bounding box indicates correct detections and the red bounding box indicate a missed detection for the baseline that was correctly detected by our proposed method.
Qualitatively, it appears that our method more reliably detects instances of cars that are small in the image, in particular in the far background, at a scale in which the pixel statistics of the image are more pronounced. Note that our method also improves performance on car detections for cases where the image is over-saturated due to increased exposure, which we are directly modeling through our proposed augmentation pipeline.
Additionally, our method produces improved detections for other effects that obscure the presence of a car, such as occlusion and shadows, even though we do not directly model these effects. This may be attributed to increased robustness to effects that lead to loss of visual information about an object in general.

\section{Conclusions}
\label{sec:concl}
We have proposed a novel sensor-based image augmentation pipeline for augmenting synthetic training data input to DNNs for the task of object detection in real urban driving scenes. Our augmentation pipeline models a range of physically-realistic sensor effects that occur throughout the image formation and post-processing pipeline. These effects were chosen as they lead to loss of information or distortion of a scene, which degrades network performance on learned visual tasks. By training on our augmented datasets, we can effectively increase dataset size and variation in the sensor domain, without the need for further labeling, in order to improve robustness and generalizability of resulting object detection networks. We achieve significantly improved performance across a range of benchmark synthetic vehicle datasets, independent of the level of photorealism. Overall, our results reveal insight into the importance of modeling sensor effects for the specific problem of training on synthetic data and testing on real data.

%
\subsection*{Acknowledgements}
This work was supported by a grant from Ford Motor Company via the Ford-UM Alliance under award N022884, and  by the National Science Foundation under Grant No. 1452793.


\newpage

\subsection*{Supplemental Material: Augmenting Real Data with Sensor Effects \footnote{Please note that these results originally appeared in the main body of a previous version of this paper, entitled \textit{Modeling Camera Effects to Improve Deep Vision for Real and Synthetic Data}.} }

Two of the most popular vehicle benchmark datasets, KITTI~\cite{kittiobject}~\cite{KITTI13} and Cityscapes~\cite{Cityscapes}, share many spatial and environmental visual features: both are captured during similar times of day, in similar weather conditions, and in cities regionally close together, with the camera located on a car pointing at the road. In spite of these similarities, images from these datasets are visibly different, a side-by-side comparison of which can be found in the upper row of example images in Figure~\ref{fig:datasets}. This suggests that these two real datasets differ in their global pixel statistics. We extend our experiments to evaluate how randomizing in the sensor effect domain can reduce the observed domain gap between real datasets. To evaluate our method in the real domain, we augment two benchmark vehicle datasets, KITTI~\cite{kittiobject}~\cite{KITTI13} and Cityscapes~\cite{Cityscapes} using the proposed method, and repeat the above experiments using Faster-RCNN. \\ \\
\noindent\textbf{S.1 Performance on Baseline Object Detection Benchmarks}\\

\setlength{\tabcolsep}{3.0pt}
\begin{table*}
\caption{Object detection trained on real data, tested on KITTI}
\label{tab:objdetreal}
\begin{center}
\begin{tabular}{cc}

\begin{tabular}{llll}
\hline\noalign{\smallskip}
\multicolumn{4}{c}{KITTI}  \\
\hline\noalign{\smallskip}
Training Set && $AP_{car}$ & Gain  \\
\noalign{\smallskip}
\hline
\noalign{\smallskip}
2975 Baseline && 79.12 & \textemdash \\
2975 Prop. Method && 81.89 &  $\uparrow$ 2.89 \\
\hline\noalign{\smallskip}
Full Baseline (6K) && 82.23 & \textemdash \\
Full Prop. Method && 83.28 & $\uparrow$ 1.05  \\
\noalign{\smallskip}\hline
\end{tabular}
&
\begin{tabular}{llll}
\hline\noalign{\smallskip}
\multicolumn{4}{c}{Cityscapes}  \\
\hline\noalign{\smallskip}
Training Set && $AP_{car}$ & Gain  \\
\noalign{\smallskip}
\hline
\noalign{\smallskip}
2975 Baseline &&  62.69 & \textemdash\\
2975 Prop. Method && 64.93 & $\uparrow$ 2.24\\
\hline\noalign{\smallskip}
\multicolumn{4}{c}{\textemdash}\\
\multicolumn{4}{c}{\textemdash}\\
\noalign{\smallskip}\hline
\end{tabular}


\end{tabular}
\end{center}
\end{table*}
\setlength{\tabcolsep}{1.4pt}
Table~\ref{tab:objdetreal} shows results trained on real data (KITTI and Cityscapes). For both real training datasets, images augmented with the proposed method see performance gains over the baseline (unaugmented) datasets. However, in general, the $AP_{car}$ gain is higher when augmenting synthetic data with sensor effects compared to augmenting real data. This is expected as, in general, synthetic data does not realistically model sensor effects such as noise, blur, and chromatic aberration as accurately as our proposed approach. \\\\ 
\noindent\textbf{S.2 Comparison to Other Augmentation Techniques}\\
\setlength{\tabcolsep}{1.0pt}
\begin{table*}
\begin{center}
\caption{We provide the results of training Faster-RCNN networks on Cityscapes augmented with various augmentation methods and tested on KITTI.}
\label{tab:otheraugs_real}
\begin{tabular}{lllllllllll}
\hline\noalign{\smallskip}

\multicolumn{10}{c}{Cityscapes}  \\
\hline
\noalign{\smallskip}
Augmentation Method           &&&&& $AP_{Car}$ &&&&& \footnotesize{Gain} \\
\noalign{\smallskip}
\hline
\noalign{\smallskip}
Baseline                   &&&&& 62.69          &&&&& \textemdash\\  
\textbf{Prop. Method}      &&&&& \textbf{64.93} &&&&& $\uparrow$ \textbf{2.24}\\  
\textit{Krishevsky et. al \cite{krizhevsky2012imagenet}}    &&&&& 63.48          &&&&& $\uparrow$ 0.79\\  
\textit{Ronneberger et. al \cite{ronneberger2015u}}   &&&&& 63.52          &&&&& $\uparrow$ 0.83\\  
\multirow{2}{10em}{\small{Random Rotation, Scale, Transl., Crop}} &&&&& 63.74 &&&&& $\uparrow$ 1.05\\
\noalign{\smallskip}\\
\hline
\end{tabular}
\end{center}
\end{table*}
\setlength{\tabcolsep}{1.4pt}
We provide the results of training Faster-RCNN networks on the full Cityscapes augmented with the above methods in Table \ref{tab:otheraugs_real}. All networks were tested on the same held-out set of KITTI images as used in the previous object detection experiments.
Our results show that our proposed method outperforms the other augmentation techniques. \\\\
\noindent \textbf{S.3 Ablation Study}\\

\setlength{\tabcolsep}{1.0pt}
\begin{table*}
\begin{center}
\caption{Ablation study for object detection trained on real data, tested on KITTI}
\label{fig:ablation_real}
\begin{tabular}{lllll lllll llll l}
\hline\noalign{\smallskip}
\multicolumn{15}{c}{KITTI}    \\
\hline
\noalign{\smallskip}
Training Set &&&&& Augmentation Type &&&&& $AP_{car}$ &&&& Gain  \\
\noalign{\smallskip}
\hline
\noalign{\smallskip}
2975 Baseline     &&&&& \it{None}    &&&&& 79.12 &&&& \textemdash \\
2975 Prop. Method &&&&& Chrom. Ab.   &&&&& 79.72 &&&& $\uparrow$ 0.60 \\
2975 Prop. Method &&&&& Blur         &&&&& 79.97 &&&& $\uparrow$ 0.85 \\
2975 Prop. Method &&&&& Exposure     &&&&& 80.35 &&&& $\uparrow$ 1.23 \\
2975 Prop. Method &&&&& Sensor Noise &&&&& 80.74 &&&& $\uparrow$ 1.62 \\
2975 Prop. Method &&&&& Color Shift  &&&&& 80.23 &&&& $\uparrow$ 1.11 \\
\noalign{\smallskip}
\hline
\hline
\noalign{\smallskip}

\multicolumn{15}{c}{Cityscapes}  \\
\hline
\noalign{\smallskip}
Training Set &&&&& Augmentation Type &&&&& $AP_{car}$ &&&& Gain  \\
\noalign{\smallskip}
\hline
\noalign{\smallskip}
2975 Baseline     &&&&& \it{None}    &&&&& 62.69 &&&& \textemdash \\
2975 Prop. Method &&&&& Chrom. Ab.   &&&&& 63.25 &&&& $\uparrow$ 0.56 \\
2975 Prop. Method &&&&& Blur         &&&&& 62.10 &&&& $\downarrow$ 0.59 \\
2975 Prop. Method &&&&& Exposure     &&&&& 62.28 &&&& $\downarrow$ 0.41 \\
2975 Prop. Method &&&&& Sensor Noise &&&&& 62.81 &&&& $\uparrow$ 0.12 \\
2975 Prop. Method &&&&& Color Shift  &&&&& 64.14 &&&& $\uparrow$ 1.45 \\
\hline
\end{tabular}
\end{center}
\end{table*}
\setlength{\tabcolsep}{1.4pt}
We trained Faster-RCNN on each of the real datasets augmented with single augmentation functions. Table ~\ref{fig:ablation_real} shows the resulting $AP_{car}$ for each network trained on real data, respectively. 
Performance increases across all ablation experiments for training. This further validates our hypothesis that sensor effects are important for closing the domain gap between real datasets. For ablation experiments trained on real data, in general, the performance increases, although blur and exposure show slight decreases in performance for networks trained on Cityscapes and tested on KITTI. Note that for augmenting Cityscapes, the color shift contributes the most to improving performance. As mentioned above, baseline Cityscapes has a noticeably distinct color cast from KITTI, and the color cast across the Cityscapes dataset is less varied in general. This result demonstrates that increasing variation in color cast across the Cityscapes dataset has a relatively high impact on cross-dataset generalization.

\end{document}